\documentclass{article}

\PassOptionsToPackage{numbers, compress}{natbib}
\usepackage[final]{Mobin.Shariq.Voice.Conversion.2016}

\usepackage[utf8]{inputenc} 
\usepackage[T1]{fontenc}    
\usepackage{url}            
\usepackage{booktabs}       
\usepackage{amsfonts}       
\usepackage{nicefrac}       
\usepackage{microtype}      
\usepackage{amsmath}
\usepackage{graphicx}
\usepackage{subfig}

\title{Voice Conversion using Convolutional Neural Networks}

\author{
  Shariq A. Mobin \\
  Redwood Center for Theoretical Neuroscience \\
  UC Berkeley \\
  \texttt{shariq.mobin@gmail.com} \\
  \And
  Joan Bruna \\
  Department of Statistics \\
  UC Berkeley \\
  \texttt{joan.bruna@gmail.com}
}

\begin{document}

\maketitle

\begin{abstract}
    The human auditory system is able to distinguish the vocal source of thousands of speakers, yet not much is known about what features the auditory system uses to do this. Fourier Transforms are capable of capturing the pitch and harmonic structure of the speaker but this alone proves insufficient at identifying speakers uniquely. The remaining structure, often referred to as timbre, is critical to identifying speakers but we understood little about it. In this paper we use recent advances in neural networks in order to manipulate the voice of one speaker into another by transforming not only the pitch of the speaker, but the timbre. We review generative models built with neural networks as well as architectures for creating neural networks that learn analogies. Our preliminary results converting voices from one speaker to another are encouraging.
\end{abstract}

\section{Introduction}
When audiologists describe what makes the sound of one person to the next sound different they first refer to the pitches of the speakers and second to the timbre of speakers. While pitch is well described by the harmonic structure the timbre is described broadly as everything besides pitch and intensity. Often, sounds with identical pitches can sound completely different. For example, we can tell the difference between a trumpet and piano playing the same pitch, this is because the timbre of each instrument gives rise to different perceptions of the sound.

One can think about vocal signals as being an entanglement of two factors - what the speaker is saying and who is saying it. The vocal signal is a non-stationary process which causes the disentanglement of these two factors to be very difficult. In this paper we will explore if it possible to hold one of these two factors constant and alternate the other. That is, we will see if it is possible to convert the speaker of a vocal signal while holding the spoken word constant. In \cite{zotkin2003pitch} it was shown that, using auditory representations inspired by the brain, it was possible to interpolate sound along the timbre axis between a trumpet and a piano. However, the model was hand engineered, in this work we seek to learn the transformational operator using neural networks.

\section{Background}
\subsection{Constant Q-Transform}
    In theory, we could train our neural network with the raw audio waveformm as input. However, in the signal processing community some type of frequency analysis of the waveform is often analyzed as this transformation makes explicit the harmonic structure of the signal. Here, we apply a constant-Q wavelet transformation (CQT) to the audio signal. This transformation has a number of desirable properties, the most important are:
\begin{enumerate}
\item The transformation uses logarithmic scaling in frequency. This is very useful when the sound wafeform spans many octaves as in the case of waveforms from the human vocal system.
\item The CQT transformation has very high temporal resolution and low spectral resolution for high frequencies whereas the converse is true for low frequncies. This is very similar to the transformation the basilar membrane of the cochlea performs on the sound waveform.
\end{enumerate}

\subsection{Deep Visual Analogy Making}
\begin{figure}[h]
    \centering
    \includegraphics[scale=0.40]{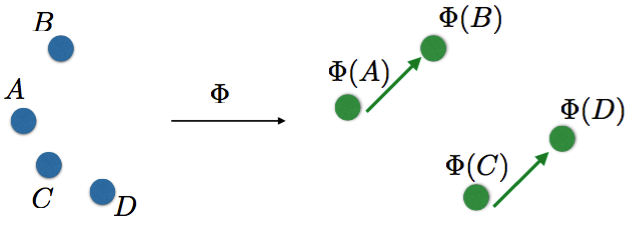}
    \caption{By learning an embedding operator, $\Phi$, we are able to linearize the analogy "A is to B as C is to D"}
    \label{embed_linear}
\end{figure}

\begin{figure}[h]
    \centering
    \includegraphics[scale=0.40]{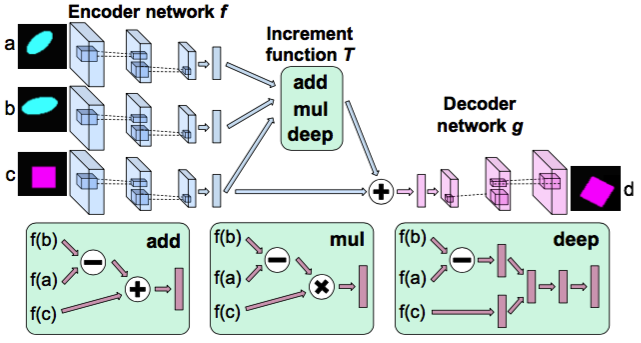}
    \caption{A visualization of the Convolutional Neural Network used in the Visual Analogy Network}
    \label{analogy_network}
\end{figure}

    Deep Visual Analogy Networks \cite{reed2015deep}  are a recent neural network architecture that has been able to achieve incredible results rotating sprites in the image domain. The goal of the network is to make analogies: "A is to B as C is to D". That is, given A, B, and C as input we would like to predict D. An example would be: "groom is to bride as king is to queen". The approach taken by this model is to learn an embedding of the input such that solving these analogies is easy, e.g. linear:
\begin{eqnarray*}
\Phi(D) - \Phi(C) \approx \Phi(B) - \Phi(A)
\end{eqnarray*}
This embedding can be visualized in Figure \ref{embed_linear}. In practice the relationship does not have to be linear, the relationship can further be approximated by more neural network layers, as in the case of our model. A visualization of the network can be seen in Figure \ref{analogy_network}.

Here our objective function is:
\begin{eqnarray*}
E = \sum_{(a,b,c,d)} \frac{1}{2}||d - g(\Phi(b) - \Phi(a) + \Phi(c)||^2
\end{eqnarray*}

\subsection{Generative Adversarial Networks}
\begin{figure}[h]
    \centering
    \includegraphics[scale=0.40]{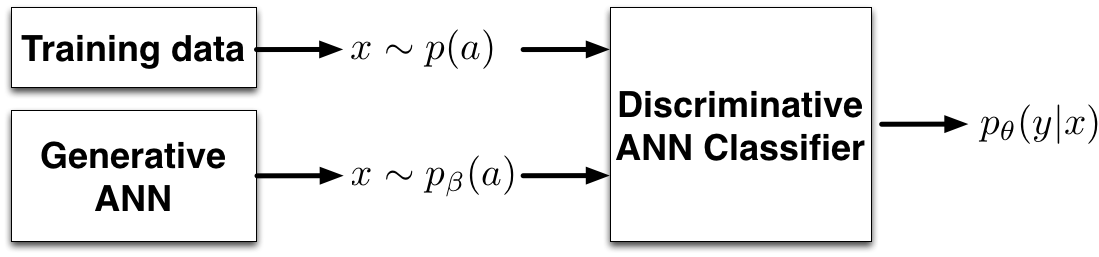}
    \caption{A visualization of the the Generative Adversarial Network idea}
    \label{adversarial_network}
\end{figure}

Generative adversarial networks (GANs) \cite{goodfellow2014generative} are a recent neural network architecture that allow for very good generative models. These networks have been used in the image domain to create very convincing images of a variety of objects \cite{denton2015deep}. The basic idea is to use one neural network that is a generator and use another neural network as a discriminator. The networks are adversarial in the sense that the generative model is trying to imitate the distribution of some true distribution, e.g. images, while the discriminative network is trying to classify images as coming from the true distribuitoin or the generative, fake, distribution. This is articulated in Figure \ref{adversarial_network}.

The goal is then to solve a minimax problem:
\begin{eqnarray*}
\underset{\beta}{\text{min}} \hspace{1mm} \underset{\theta}{\text{max}} \hspace{1mm} \Big[ \mathbb{E}_{x \sim p(a)} \text{log} \hspace{1mm} p_{\theta} (y=\text{`real’}|x) + \mathbb{E}_{x \sim p_{\beta}(a)} \text{log} \hspace{1mm} p_{\theta} (y=\text{`fake’}|x) \Big]
\end{eqnarray*}

In practice optimizing these networks is very difficult and realizes on many tricks.

\section{Model}
Here, we combine the ideas from Deep Visual Analogy Networks (VANs) and Generative Adversarial Networks (GANs) in order to create a model capable of doing voice conversion. A VAN serves as our generative model in the GAN sense. The discriminator of our GAN is then implemented by a classifier which distinguishes not only real and fake CQT samples but what speaker and word category the sample belongs to. This can be summarized by the new minimax equation:

\begin{eqnarray*}
\underset{\beta}{\text{min}} \hspace{1mm} \underset{\theta}{\text{max}} \hspace{1mm} \Bigg[ \sum_{(w,s)} &\mathbb{E}_{x \sim p(a|W=w,S=s)} \text{log} \hspace{1mm} p_{\theta} (W=w, S=s|x) \hspace{1mm} + \\
&\mathbb{E}_{x \sim p_{\beta}(a|W=w,S=s)} \text{log} \hspace{1mm} p_{\theta} (w=\text{`fake'}, s=\text{`fake'}|x) \Bigg]
\end{eqnarray*}

Note, in order to weight the classifier towards distinguishing fake words and speakers we bias half of the samples in a batch to be from the generative model. The sampling is otherwise uniform over the speaker and word. Our code can be viewed at https://github.com/ShariqM/smcnn. The model parameters are found in models/cnn.lua.

\subsection{Results}

\begin{figure}[h]
    \centering
    \includegraphics[scale=0.40]{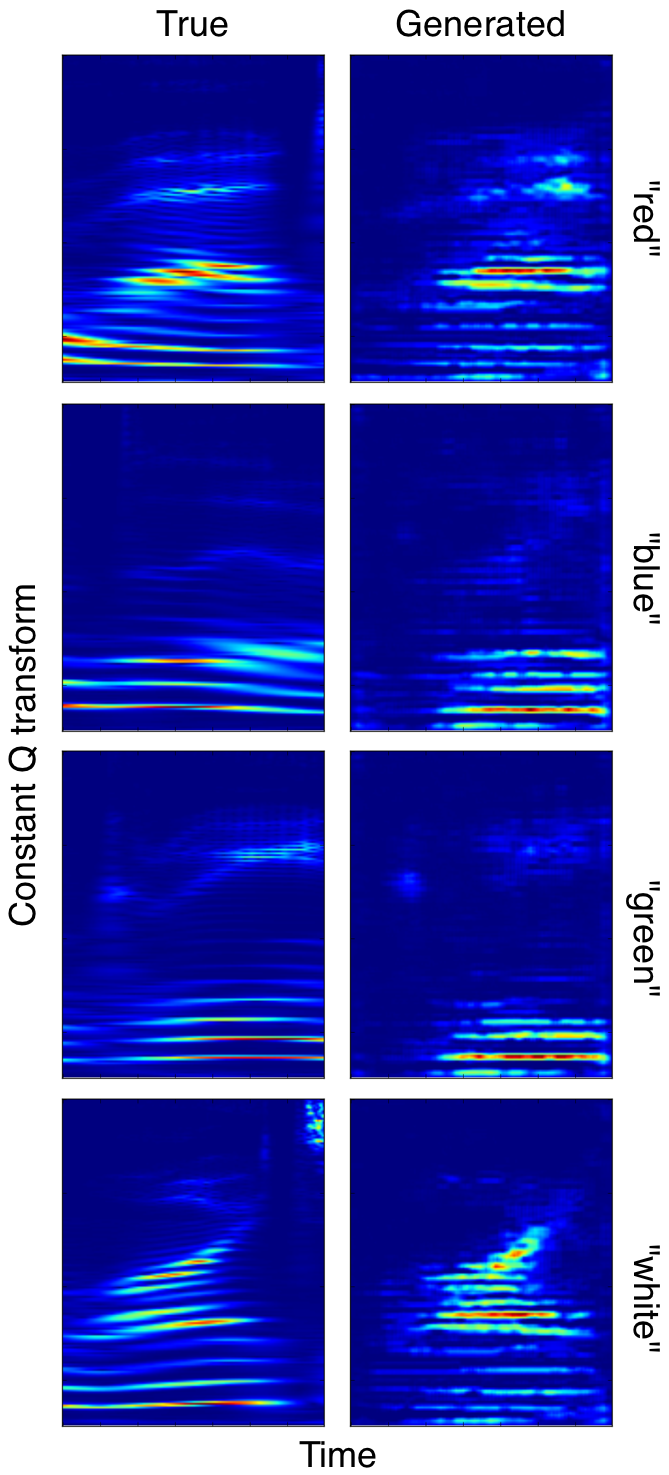}
    \caption{The left column corresponds to samples of Speaker 2 saying the color indicated on the row from the training data. The right column corresponds to generates samples from our model of the same speaker and color.}
    \label{results}
\end{figure}

Our results can be seen in Figure 4. While the model is able to capture the harmonic structure of the speaker well, the frequency resolution is a bit poor. This likely an artifact of up sampling in the decoding phase. Note that this data is from the training set and there is only 1 speaker and 4 words in the data set. The audio samples can be heard at the following links:

\begin{itemize}
\item  https://dl.dropboxusercontent.com/u/7518467/Bruna/model2/GAN\_results/results\_red.wav
\item https://dl.dropboxusercontent.com/u/7518467/Bruna/model2/GAN\_results/results\_blue.wav
\item https://dl.dropboxusercontent.com/u/7518467/Bruna/model2/GAN\_results/results\_green.wav
\item https://dl.dropboxusercontent.com/u/7518467/Bruna/model2/GAN\_results/results\_white.wav
\end{itemize}
A sample from the true distribution comes first, and then one from our generative model, for each file.

\subsection{Conclusion}
We began by developing algorithms in order to transfer the timbre of one speaker to another. Our algorithms were able to produce speech that occassionally sounded perceptually similar to the target speaker but work remains to be done. Training Generative Adversarial Networks has proven very difficult in practice and more time will need to be spent understanding how best optimize the Conditional Generative Adversarial Network model developed here.

{\small
\clearpage
\bibliographystyle{plain}
\bibliography{references.bib}}

\end{document}